\documentclass[letterpaper, 10 pt, conference]{ieeeconf}  

\IEEEoverridecommandlockouts                              

\usepackage{cite}   
\usepackage{graphics}
\usepackage{epsfig}
\usepackage{times} 
\usepackage{amsmath} 
\usepackage{amssymb}  
\usepackage{graphicx}
\usepackage{bbding}
\usepackage{balance}
\usepackage{lipsum}

\usepackage{xcolor}
\usepackage{booktabs} 
\usepackage{bm}
\usepackage{algorithmic}
\usepackage{textcomp}
\usepackage{url}
\usepackage{silence}
\usepackage{tabularx}
\usepackage{makecell}
\usepackage{amsmath}
\usepackage{amsfonts}
\usepackage{multirow}
\usepackage{bbding}
\usepackage{threeparttable}

\usepackage{array}
\usepackage{stfloats}
\usepackage{float}
\newcolumntype{C}[1]{>{\centering\arraybackslash}p{#1}}
\usepackage{subcaption}
\definecolor{customgreen1}{HTML}{9BCD9B}

\begin{document}

\title{\LARGE \bf
MUJICA: Multi-skill Unified Joint Integration of Control Architecture for Wheeled-Legged Robots
}
\author{\small Yuqi Li$^{1}$, Peng Zhai$^{1,*}$, Yueqi Zhang$^{1}$, Xiaoyi Wei$^{1}$, Quancheng Qian$^{1}$, Zhengxu He$^{2}$, Qianxiang Yu$^{2}$ and Lihua Zhang$^{1,*}$
\thanks{$^{1}$Yuqi Li, Yueqi Zhang, Xiaoyi Wei, Quancheng Qian, Peng Zhai, and Lihua Zhang are with the College of Intelligent Robotics and Advanced Manufacturing, Fudan University, Shanghai 200433, China
        {\tt\small \{yuqili24, zhangyq23, weixy23, qcqian24\}@m.fudan.edu.cn; \{pzhai, lihuazhang\}@fudan.edu.cn}}%
\thanks{$^{2}$Zhengxu He, and Qianxiang Yu are with Power China Huadong Engineering Corporation Limited, Hangzhou, China
        {\tt\small \{he\_zx, yu\_qxl\}@hdec.com}}%
\thanks{* Corresponding Author.}
\\
{Project Page: \url{https://hyzenthlayer.github.io/mujica/}}
}
\maketitle

\thispagestyle{empty}
\pagestyle{empty}
\begin{abstract}
Wheeled-legged robots hold promise for traversing complex terrains and offer superior mobility compared to legged robots. However, wheeled-legged robots must effectively balance both wheeled driving and legged control. Furthermore, due to noisy proprioceptive sensing and real-world motor constraints, realizing robust and adaptive locomotion at peak performance of motors remains challenging. We propose the Multi-skill Unified Joint Integration of Control Architecture (MUJICA), a unified, fully proprioceptive control framework for wheeled-legged robots that  integrates diverse low-level skills—including omnidirectional moving, high platform climbing, and fall recovery—within a single policy. All skills, distinguished by unique indicator variables, are trained jointly with accurate DC-motor constraint modeling. Additionally, a high-level skill selector is learned to dynamically choose the optimal skill based solely on proprioceptions, enabling adaptive responses to the surrounding environment. Therefore, MUJICA enhances sim-to-real robustness and enables seamless transitions across diverse locomotion modes, facilitating autonomous adjustment to the environment. We validate our framework in both simulation and real-world experiments on the Unitree Go2-W robot, demonstrating significant improvements in adaptability and task success in unstructured environments.
\end{abstract}

\section{INTRODUCTION}
\label{sec:introduction}

With significant advances in motor torque and computational power of hardware, legged robots have achieved impressive progress in traversing challenging environments. Their ability to adapt to uneven terrains and recover from perturbations makes them suitable for real-world tasks such as inspection and disaster response \cite{tang2025deep}. Despite these advances, designing control algorithms for legged robots remains challenging. Controllers must simultaneously achieve motion diversity, adaptability to different terrains, and safe deployment under real-world actuation constraints. A variety of approaches have been proposed and have demonstrated robustness in tasks such as stair climbing, slope traversal, and perturbation recovery \cite{huang2025moe,zhang2024deep,nahrendra2023dreamwaq,lu2025fr,11155164}. However, owing to inherent limitations of the robot's structure, legged locomotion is often limited in efficiency and speed on flat ground, and lacks the capability of traversing certain challenging terrains, such as high platforms.

Wheeled-legged robots offer a promising alternative by combining the efficiency of wheels with the legs' ability to cross obstacles. This hybrid design enables rapid traversal on smooth surfaces while preserving the ability to handle rough terrains and obstacles \cite{lee2024learning}. 
Nevertheless, the full potential of wheeled-legged robots in complex real-world environments has yet to be realized because most existing control frameworks for wheeled-legged robots are modified from legged locomotion methods \cite{lee2024learning,chamorro2024reinforcement}, and thus inherit their limitations.

\begin{figure}
    \centering
    \captionsetup{font=footnotesize}
    \includegraphics[width=\linewidth]{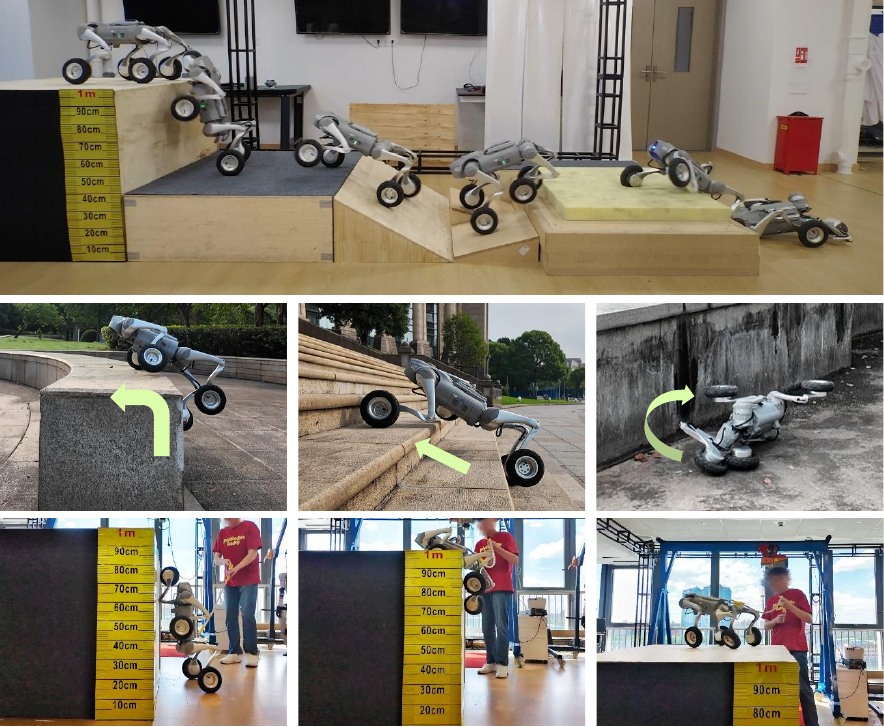}
    \caption{Real-world demonstrations of MUJICA on the wheeled-legged robot. \textbf{Top row}: automatic skill switching during continuous locomotion. \textbf{Middle row}: execution of three representative skills, including high platform climbing, stair ascent, and fall recovery. \textbf{Bottom row}: snapshots of the robot climbing onto a 1-m-high platform. The results highlight MUJICA’s ability to perform diverse skills and seamlessly transition between them without external perception.}
    \label{fig:head_pic}
\end{figure}

We identify three key limitations in current legged and wheeled-legged locomotion approaches. First, most multi-skill blind controllers based solely on proprioception can only handle either a narrow range of similar skills with limited motion diversity, such as walking on flat or inclined terrains or skills with relatively low difficulty, such as climbing stairs and slopes \cite{huang2025moe,shah2023mtac,zhang2024deep,nahrendra2023dreamwaq}. Second, for handling dissimilar skills with distinct dynamics, such as galloping, ambling and crawling, controllers often rely on manual switching\cite{bellegarda2025allgaits,li2023versatile}. Moreover, these methods are primarily designed to showcase multi-skill learning paradigms, rather than to adapt to diverse and complex terrains.
Third, in order to ensure safe deployment in the real world, limitations on robot joints are of vital importance. In the context of reinforcement learning (RL), these physical limitations are often transformed into action penalties or incorporated as constraints in optimization problems\cite{zhang2022penalized,kim2024not,gangapurwala2020guided}.  However, most learning-based methods apply simplified actuation constraints, ignoring the velocity-and-position-dependent torque limitations of direct current (DC) motors. Methods like \cite{yang2024agile} simply model the relationship between desired torques and measured torques. The lack of precise motor constraints limits both the exploitation of motor performance and the safety of sim-to-real transfer. Consequently, these issues limit the performance and safety of wheeled-legged locomotion on challenging terrains.

To address these issues, we present the Multi-skill Unified Joint Integration of Control Architecture (MUJICA) for wheeled-legged robots, a proprioception-driven control framework that unifies multiple locomotion skills—including omnidirectional moving, high platform climbing, and fall recovery—within a single network. 
A central feature of MUJICA is its emphasis on safety-aware learning and reliable sim-to-real transfer. We explicitly model the relationship between motor velocity and maximum torque as hard DC-motor constraints to prevent unsafe behaviors during deployment and allow the robot to exploit its motors more effectively during challenging maneuvers. Furthermore, we incorporate a high-level selector that leverages proprioceptive feedback to automate skill transitions, enabling smooth and adaptive switching across tasks without external perception.

The main contributions of this work are as follows:
\begin{itemize}
\item We propose a unified control architecture that jointly learns diverse and challenging skills, including omnidirectional locomotion, high platform climbing, and fall recovery, within a single blind policy.
\item We present a safety-aware learning framework based on hard DC-motor constraints to ensure successful sim-to-real transfer.
\item We train a high-level skill selector to automate skill transfer based on proprioceptions and the unified velocity tracking rewards.
\item We validate the effectiveness of MUJICA through extensive simulation and real-world experiments on the Unitree Go2-W robot.
\end{itemize}

\section{RELATED WORK}
\subsection{Learning-Based Blind Control for Legged Robots}

Since only a few studies focus on wheeled-legged locomotion and most resemble legged control, this section mainly concentrates on legged locomotion. Blind control methods rely solely on proprioceptive feedback, such as joint states and IMU data, without exteroceptive sensors. Early works demonstrate end-to-end deep RL policies for quadrupeds wandering on flat indoor surfaces \cite{tan2018sim}. Policies are trained in simulation and deployed zero-shot on real robots. To improve the performance on unstructured terrains, privileged information is widely applied during policy training. RMA \cite{kumar2021rma} leverages an adaptation module to mimic a privileged base policy, while DreamWaQ\cite{nahrendra2023dreamwaq} implements an asymmetric actor-critic to better train the policy through a critic with privileged observations. Another line of work estimates unobservable states through history observations, enabling better terrain adaptation. DreamWaQ \cite{nahrendra2023dreamwaq} and HIM \cite{long2023hybrid} both estimate the base linear velocity and a latent state vector to strengthen the robustness of locomotion, and FR-Net \cite{lu2025fr} predicts the robot collision probability and mass distribution to avoid undesirable collisions and enable deployment on diverse robots. Nevertheless, these methods do not incorporate a unified embodiment state estimation that captures diverse robot-specific information, limiting their applicability to multi-skill policy deployment.

On the other hand, policies that perform well in simulation do not necessarily guarantee safe real-world performance due to discrepancies in environment and robot dynamics. Therefore, enforcing physical constraints during training is essential to ensure safe and reliable deployment. 
Recent advances in constrained reinforcement learning have yielded several innovative approaches to address the challenge of policy optimization under constraints. To enhance computational tractability and simplify reward design, some approaches reformulate the traditionally constrained policy iteration process into an equivalent unconstrained optimization formulation \cite{zhang2022penalized,kim2024not}.  From another perspective, CaT \cite{chane2024cat} ingeniously reinterprets hard constraints as probabilistic termination mechanisms, achieving robust constraint satisfaction while maintaining computational efficiency. However, studies such as Dadiotis et al. \cite{dadiotis2025dynamic} and ALARM \cite{zhou2025alarm} only constrain the joint torques by their maximum allowable values, neglecting the fact that the maximum torque output of a DC motor decreases at high speeds, while \cite{yang2024agile} merely models the relationship between desired torques and measured torques. This oversight limits the robot's safety performance under extreme operating conditions. For wheeled-legged robots, exploiting extreme motor capabilities in coordination with wheel–leg coordination is essential to maximize performance, further highlighting the need for accurate torque modeling.

\subsection{Multitask Learning in Robotics}
The goal of multitask learning in robotics is to enable robots to accomplish different tasks or acquire various skills through a unified policy. Although  \cite{nahrendra2023dreamwaq,long2023hybrid} enable robots to traverse stairs and flat terrains, the action spaces of these skills do not differ substantially. Therefore, multitask frameworks are required to handle extensively different tasks and enable robots to learn different skills. Existing frameworks mainly fall into two types: teacher-student policy distillation and hierarchical reinforcement learning. Teacher-student policy distillation first trains multiple expert policies, each specialized in a single task, and then uses supervised learning to train a student policy to imitate these experts \cite{parisotto2015actor,teh2017distral}. Hierarchical reinforcement learning also adopts a two-stage training scheme where multiple low-level single-skill expert policies are first trained. For high-level policy, some studies apply mixture-of-expert (MOE) framework to achieve multi-skill learning such as fall recovery, quadrupedal walking, and bipedal walking, in which a gating network is trained to compute a weighted combination of these experts, yielding a hybrid policy that integrates features from various skills \cite{yang2020multi,huang2025moe,yu2025discovery}. In contrast, some methods employ a high-level policy to select only one expert policy for execution. Nevertheless, the distillation framework and the MOE framework are unable to resolve conflicts between low-level behaviors (such as conflicting actions for turning left and right) \cite{shah2023mtac,zhang2024deep}. In addition, approaches relying on training multiple separate policies often increase network complexity and reduce training efficiency. Other approaches employ imitation learning from diverse reference motions \cite{li2022versatile,li2023versatile}, but these methods mainly emphasize the  multi-skill learning paradigms, with limited attention to adaptation across diverse and challenging terrains. Chamorro et al. \cite{chamorro2024reinforcement} specialize in wheeled-legged stair climbing and adds a terrain boolean variable into the observation to distinguish stairs from other terrains. However, these two tasks are relatively easy and cannot leverage the full potential of wheeled-legged robots, and the challenge of multi-skill learning in wheeled-legged robots still remains unaddressed.

\begin{figure*}[htbp]

\centering
\captionsetup{font=footnotesize}

\includegraphics[width=\textwidth]{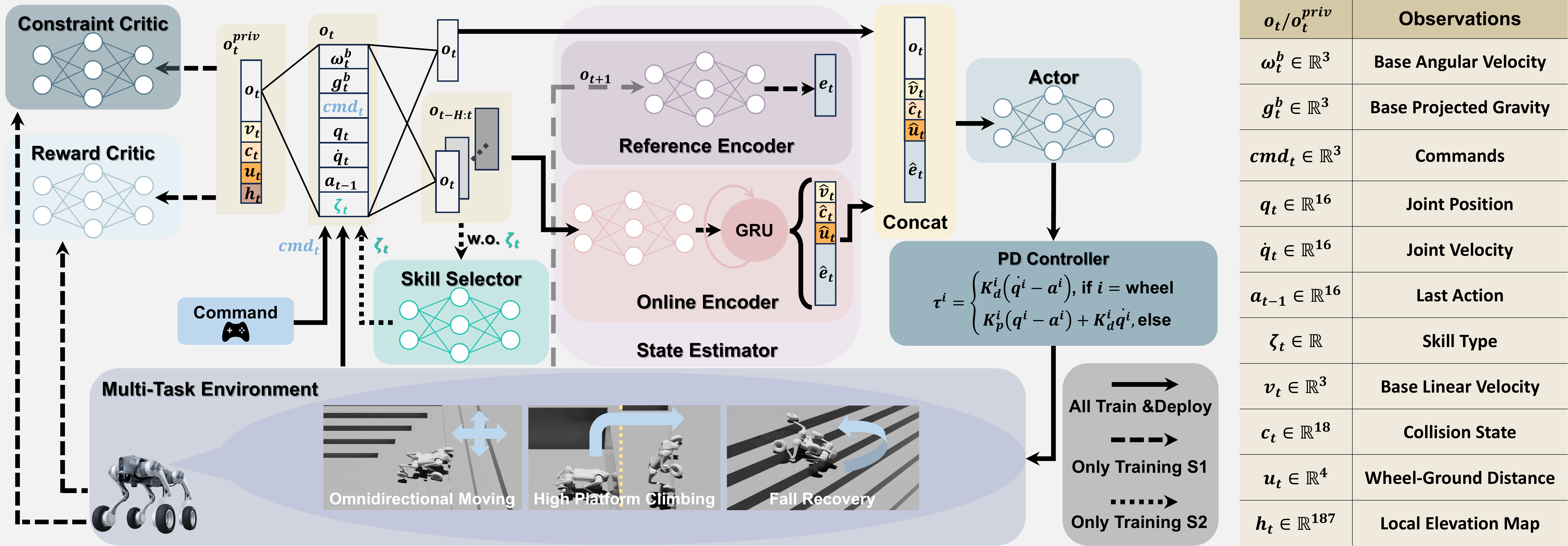}

\caption{An overview of MUJICA framework. Each task is associated with a unique skill indicator. During Training S1, the state estimator learns to predict the latent state vector, base linear velocity, wheel-ground distances and robot segment collisions from proprioceptions. The actor then takes the estimation and current observation as input and generate actions. The reward critic and the constraint critic evaluate the state value and constraint violation through privileged observations. A skill selector is applied during Training S2 to learn skill transfer, endeavoring to adapt to different terrains and states.}
\label{fig:res}

\end{figure*}
\section{Problem Formulation}
\subsection{C-POMDP Framework}
\label{3a}
Wheeled-legged robots operate in unstructured environments where full state observability is often infeasible due to sensor noise. Meanwhile, safety-critical constraints (e.g., contact force, motor limits) must be satisfied to enable reliable deployment in the real world. Therefore, we formulate the problem as a Constrained Partially Observable Markov Decision Process (C-POMDP), seeking a policy $\pi^*$ that maximizes long-term return while meeting all constraints:

\begin{equation}
\label{optimize}
    \begin{gathered}
    \max \mathbb{E}_{\pi}\left[\sum_{t=0}^\infty \gamma^t R\left(\boldsymbol{s}_{t},\bm{a}_t,\boldsymbol{s}_{t+1}\right)\right] \\
    \text{s.t.}\ \  \mathbb{E}_{\pi}\left[\sum_{t=0}^\infty \gamma^t C_i\left(\boldsymbol{s}_{t},\boldsymbol{a}_t,\boldsymbol{s}_{t+1}\right)\right] \leq \delta_i,\forall i \in \left\{1,...,k\right\} 
    \end{gathered}
\end{equation}
where $R$ represents the reward while $C_i$ denotes the $i$th constraint with corresponding limit $\delta_i$, which will be detailed in Section \ref{4c}.

\subsection{Observation Space and State Space}
\label{3b}

Since the proposed framework is designed for a blind policy, 
the policy is conditioned on an observation space consisting only of proprioceptions. Simultaneously, we define the state space as the privileged observation space including task-relevant and physically informative quantities, from which the critic networks receive input (detailed in Section \ref{sec:aac}). Thus, the observation and the privileged observation at time $t$ are defined as: 
\begin{equation}
\boldsymbol{o}_t=\left[\boldsymbol{\omega}^b_t,\boldsymbol{g}^b_t,\boldsymbol{cmd}_t,\boldsymbol{q}_t,\boldsymbol{\dot{q}}_t,\boldsymbol{a}_t,\zeta_t\right]^T
\end{equation}

\begin{equation}
    \boldsymbol{s}_t\triangleq \boldsymbol{o}^{priv}_t=\left[\boldsymbol{o}_t,\boldsymbol{v}_t,\boldsymbol{c}_t,\boldsymbol{u}_t,\boldsymbol{h}_t\right]^T
\end{equation}

\subsection{Action Space}
\label{3c}
The action space is defined as the control commands for all joint motors. For each leg joint, the action corresponds to the angular offset from the predefined default posture, while 
for the wheel joints, the action specifies the desired motor velocity.
At each timestep, the policy outputs an action vector $\boldsymbol{a}_t$, which is then converted into motor torques through a PD controller formulated as:
\begin{equation}
\tau^i_t = 
\begin{cases}
K_d^i \left( \dot{q}^i_t - a_t^i \right), & \text{if } i = \text{wheel} \\
K_p^i \left( q_t^i - a_t^i \right) - K_d^i \dot{q}_t^i, & \text{otherwise}
\end{cases}
\end{equation}
where $\tau^i_t$ is the target torque output by the $i$th motor. $K_p^i$ and $K_d^i$ separately denote stiffness and damping.

\section{Method}
In this section, we present the detailed methodology of our proposed MUJICA framework. The overall architecture is illustrated in Fig. \ref{fig:res}. 
\subsection{State Estimator}

Relying solely on proprioceptions, the robot cannot directly access terrain geometry or contact conditions, which are crucial for skill execution. Thus, to infer the nearby environment, inspired by HIM \cite{long2023hybrid}, which constructs an internal model to predict system responses to disturbances, we develop a state estimator based on a buffer of past observations. Aiming to  capture temporal dependencies, the estimator utilizes an online encoder to process the past $H=6$ frames of observations through a GRU-based network. In addition to estimating the latent state vector $\boldsymbol{\hat{e}}_{t}$ and the robot base linear velocity $\boldsymbol{\hat{v}}_t$, we also predict collision probabilities $\boldsymbol{\hat{c}}_t$ for each robot component and the wheel-ground distance $\boldsymbol{\hat{u}}_t$. Explicitly predicting velocity helps the robot comprehend the locomotion dynamics and distinguish between stable and unstable locomotion regimes. The wheel-ground distance estimation reflects terrain roughness, allowing the robot to suppress unnecessary leg lifting on flat ground and better coordinate wheel–leg interaction. Inferring collision probabilities helps the robot detect contact with obstacles, such as using head collisions to sense upcoming ledges for climbing, and enables terrain adaptive recovery behaviors.

The estimation process can be formulated as follows:
\begin{align}
\boldsymbol{f}_t=\operatorname{GRU}\left(\operatorname{NN}\left(\boldsymbol{o}_{t-H:t}\right),\boldsymbol{f}_{t-1}\right)
\end{align}
where $\boldsymbol{f}_t = [\boldsymbol{\hat{v}}_t,\boldsymbol{\hat{c}}_t,\boldsymbol{\hat{u}}_t,\boldsymbol{\hat{e}}_{t}]^T$, while $\operatorname{NN}$ indicates fully connected layers.

On the purpose of effectively training the online encoder, we introduce a reference encoder that maps the successor observation $\boldsymbol{o}_{t+1}$ to a latent state $\boldsymbol{e}_t$ as the supervision target, and a contrastive learning loss is applied.

To summarize, the estimator loss can be formulated as:
\begin{equation}
\mathcal{L}^{\text{Estimate}}=\mathcal{L}^{\text{Pred}}+\mathcal{L}^{\text{SwAV}}(\boldsymbol{e}_{t},\boldsymbol{\hat{e}}_{t})
\end{equation}
\begin{equation}
    \mathcal{L}^{\text{Pred}}=\mathcal{L}_{\text{MSE}}(\boldsymbol{v}_t,\boldsymbol{\hat{v}}_t) + \mathcal{L}_{\text{BCE}}(\boldsymbol{c}_t,\boldsymbol{\hat{c}}_t) + \mathcal{L}_{\text{MSE}}(\boldsymbol{u}_t,\boldsymbol{\hat{u}}_t)
\end{equation}

where $\mathcal{L}_{\text{MSE}}$ equals mean squared error loss, $\mathcal{L}_{\text{BCE}}$ equals binary cross-entropy loss, and SwAV is the contrastive learning loss following \cite{caron2020unsupervised}.

\subsection{Asymmetric Actor-Critic Framework}
\label{sec:aac}

As formalized in Section \ref{3b}, the wheeled-legged robot operates under partial observability during deployment, yet requires safety assessment based on unobservable terrain properties. Meanwhile, the critic exposed to full state information learns the value
function much faster, better guiding the learning of the actor\cite{pinto2017asymmetric}. Therefore, MUJICA proposes an asymmetric actor-critic framework. The policy network $\pi_{\theta}$ receives as input the history-embedded observation $\boldsymbol{f}_t$ from the online encoder and the current observation $\boldsymbol{o}_t$. 

In parallel, to address the constrained optimization problem in Equation (\ref{optimize}), we adopt the P3O \cite{zhang2022penalized} framework, which employs a reward critic 

 and a constraint critic. Both critics are 

 conditioned on privileged observations $\boldsymbol{o}_t^{priv}$ to separately evaluate the state value and the constraint violation.

P3O transforms the constrained optimization problem into an unconstrained objective using a penalty-based iterative approach.

Consequently, P3O loss can be formulated as:
\begin{equation}
    \mathcal{L}^{\text{P3O}}(\theta) = \mathcal{L}^{\text{CLIP}}_R(\theta)+\kappa \sum_{i=1}^k \max\{0,\mathcal{L}^{\text{CLIP}}_{C_i}(\theta)\}
\end{equation}
\begin{equation}
    \begin{aligned}
\mathcal{L}^{\text{CLIP}}_R(\theta) &= \underset{\substack{s \sim d^{\pi} \\ a \sim \pi}}{\mathbb{E}} \left[ -\min \{r(\theta) A^{\pi}_R(s, a),\right.\\
&\left. \text{clip}(r(\theta),1-\epsilon,1+\epsilon)A^{\pi}_R(s, a)\}\right]
    \end{aligned}
\end{equation}
\begin{equation}
    \begin{aligned}
\mathcal{L}^{\text{CLIP}}_{C_i}(\theta)=\underset{\substack{s \sim d^{\pi} \\ a \sim \pi}}{\mathbb{E}} \left[ \max \left\{ r(\theta) A^{\pi}_{C_i}(s, a),\right.\right. \text{clip}(r(\theta),\\
\left.\left.1-\epsilon,1+\epsilon )A^{\pi}_{C_i}(s, a)\right\}+ (1 - \gamma)(J_{C_i}(\pi) - \delta_i) \right] 
    \end{aligned}
\end{equation}

where $r(\theta)$ is the importance sampling ratio between the new and old policies, $A^{\pi}_R$ and $A^{\pi}_{C_i}$ are the reward and $i$th constraint advantages, $d^{\pi}(s)$ is the discounted future state distribution, and $\kappa$ is the penalty factor.
$\mathcal{L}^{\text{CLIP}}_R$ is the clipped PPO surrogate objective\cite{schulman2017proximal}, ensuring stable reward-driven updates, while $\mathcal{L}^{\text{CLIP}}_{C_i}$ encodes constraint-specific losses with clipping to prevent overly aggressive updates. $(J_{C_i}(\pi) - \delta_i)$ penalizes expected constraint costs exceeding $\delta_i$, scaled by $(1-\gamma)$. Summed over all constraints, the P3O objective maximizes reward while suppressing unsafe behaviors, yielding policies that are both performant and safe. To enable the robot to gradually acquire higher-difficulty skills, we design a curriculum learning scheme to train the actor–critic network (see Section \ref{sec:curriculum} for details).

\begin{figure}[htbp]
  \centering
  \captionsetup{font=footnotesize}
    \includegraphics[width=\linewidth]{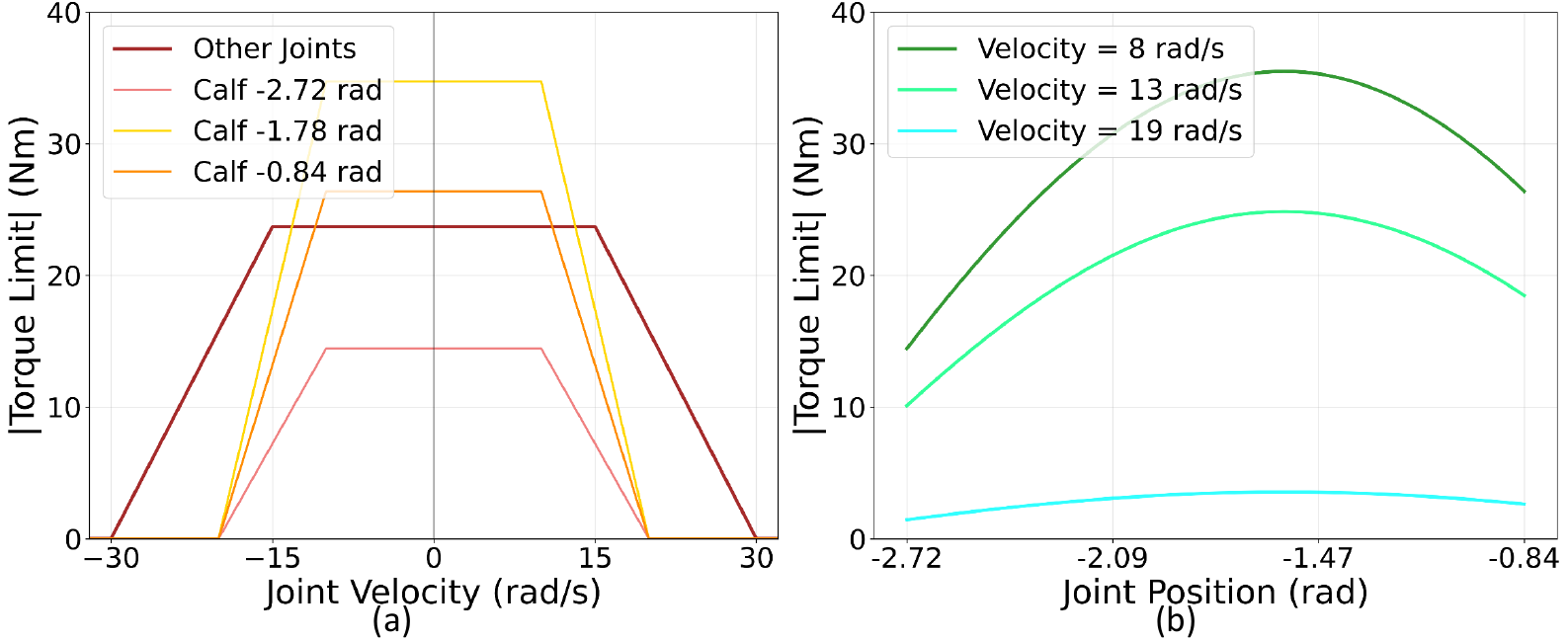}
  \caption{(a) The velocity-dependent torque limit of all joints. Calf joint limits are also influenced by joint position. (b) The position-based torque limit of calf joints.}
  \label{fig:motor}
\end{figure}
\subsection{Multi-Task Reward and Constraint}
\label{4c}
On the purpose of learning various skills within a single policy, we propose the skill indicator $\zeta_t$. Each skill is associated with a unique indicator variable so that the observation spaces of individual skills are disentangled along this specific dimension. Consequently, through the skill indicator $\zeta_t$, our proposed architecture can be extended to various wheeled-legged tasks. In this study, we design three representative and challenging tasks: (i) omnidirectional moving on ordinary terrains, (ii) high platform climbing, and (iii) fall recovery. The omnidirectional locomotion skill requires tracking linear and angular velocities, while the high platform climbing skill only needs to track linear velocity. For fall recovery, the robot needs to learn to flip itself over from arbitrary poses. 

\begin{table}[htbp]
\centering
\captionsetup{font=footnotesize}
\caption{Primary rewards and constraints of MUJICA}
\label{tab:reward}
\renewcommand\arraystretch{1.5}
\begin{center}
\begin{tabular}{c|c|c}
\hline
\makecell[c]{Reward $R_t$\\or Constraint $C_t$} & Formula& \makecell[c]{Applicable\\ Tasks} \\ 
\hline
$R_{cmd_v,t}$ & $\exp\left(-\lVert\boldsymbol{cmd}_{xy,t}-\boldsymbol{v}_{xy,t}\rVert^2/\sigma^2\right)$ & i,ii\\
\hline
$R_{cmd_{\omega},t}$ & $\exp\left(-\lVert\boldsymbol{cmd}_{\omega,t}-\omega_{z,t}\rVert^2/\sigma^2\right)$ & i\\
\hline
$R_{gravity,t}$ & $\exp\left(-\angle\left(\boldsymbol{g}^b_t,\boldsymbol{g}^{world}\right)/\sigma^2\right)$  & iii\\
\hline
$R_{poserr,t}$ & \makecell[c]{$\exp\left(-\lVert\boldsymbol{q}-\boldsymbol{q}_{stand}\rVert^2)\right/\sigma^2)$ \\ if $|\angle\left(\boldsymbol{g}^b_t,\boldsymbol{g}^{world}\right)| < \epsilon$}  & iii\\
\hline
\hline
$C_{DC-motor,t}$ & $\sum_{i=1}^{16} \mathbf{1}_{|\tau^i_t| \geq \tau_{\text{limit}}^i}$  & i,ii,iii\\
\hline
$C_{collision,t}$ & $\sum_i c_t^i, i=\text{thigh,calf}$ & i,ii \\
\hline
\end{tabular}
\end{center}
\end{table}

The primary and task-relevant rewards and constraints are listed in TABLE \ref{tab:reward}, while others are supplemented in the video.
Notably, collision constraints are imposed on the thigh and calf to encourage the robot to interact with the environment using wheels rather than legs. 
Moreover, to account for the physical limitations of electric motors, we introduce the DC-motor constraint to all the tasks. Specifically, inspired by the motor operating regions from \cite{shin2023actuator}, shown in Fig. \ref{fig:motor}, each actuator is subject to a velocity-dependent torque limit, where the maximum torque remains constant at low speeds but decreases linearly with increasing velocity at high speeds. The calf joints connect the thigh and calf links, so the torque is influenced by both joint velocity and joint position, with the maximum torque exhibiting a cosine dependence on the joint position at fixed velocity, whereas other joint motors are solely affected by joint velocity. The specific values are from the official motor manual of Unitree.

\begin{figure*}[htbp]
    \centering
    \captionsetup{font=footnotesize,justification=raggedright,singlelinecheck=false}
    \includegraphics[width=\textwidth]{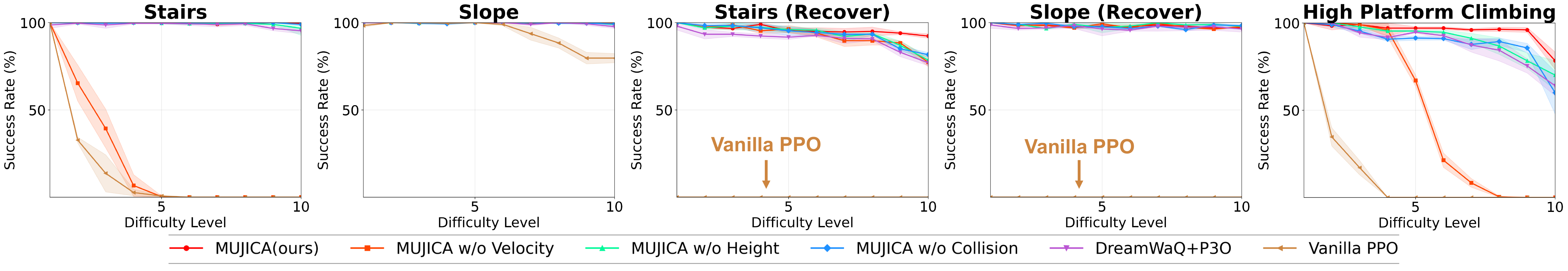}
    \caption{Success rate of different algorithms on all difficulty levels of representative tasks.}
\label{fig:compare}

\end{figure*}
\subsection{Skill Selector}
After training several robust locomotion skills (Training S1), to enable adaptive behavior across diverse terrains, we adopt a proprioception-based high-level skill selector that automates the activation of low-level locomotion policies, i.e., chooses the appropriate $\zeta_t$ automatically. In particular, the high-level selector takes the last $H$ observations $\boldsymbol{o}_{t-H:t}$, excluding $\zeta_t$, as input, and outputs a probability distribution over the predefined set of locomotion skills. An asymmetric actor-critic network is implemented to train the skill selector on top of the frozen skill policies through the unified velocity tracking rewards (Training S2).
This hierarchical design simplifies low-level learning and improves generalization through flexible skill composition.

\section{EXPERIMENTS}
We train MUJICA on diverse terrains and conduct extensive experiments in both simulation and the real world to validate its effectiveness and peak performance. In detail, the following experiments aim to address:
\begin{itemize}
\item Can MUJICA effectively complete diverse tasks, and how does its performance compare with baseline methods and ablated variants? (In Section V-B)
\item Does the proposed skill selector enable reliable switching across tasks with large variations in characteristics, leading to improved sequential-task success rates? (In Section V-C)
\item Can MUJICA policies be transferred to real wheeled-legged robots, demonstrating zero-shot sim-to-real performance? (In Section V-D)
\end{itemize}
\subsection{Training}
\label{sec:curriculum}
\subsubsection{Settings}

All experiments in simulation are conducted using the IsaacLab platform with 4,096 parallel environments on a single NVIDIA RTX 4090 GPU. We choose the Unitree Go2-W shown in Fig. \ref{fig:res} to validate the proposed architecture. The low-level controller and the high-level selector both work at a frequency of 50 Hz, while the physics simulation is executed at 200 Hz. The low-level training process is carried out for 30,000 iterations, and the high-level training is conducted for 10,000 iterations. In the fall recovery task, each episode lasts 6 s, including 2 s of free fall from the air with a random posture and 4 s for recovery, during which the task will not terminate. For the other tasks, each episode lasts 20 s and terminates when the base collision occurs.

\subsubsection{Curriculum Learning}
To enable the robot to gradually explore motions that adapt to challenging terrains, following \cite{rudin2022learning}, we apply a curriculum learning framework. 

We construct a $33\times20$ grid-shaped multi-task environment, in which 4,096 robots are assigned to different rows, each corresponding to a specific terrain and its corresponding task, while each column represents a curriculum difficulty level.

During Training S1, we explicitly set the correct skill indicator according to the task assigned to each robot. For both omnidirectional locomotion and fall recovery, the terrains are set as stairs, slopes, discretized terrains and rough terrains. For high platform climbing, the terrain is set as a large depression on the ground that requires the robot to climb out and enables climbing in any direction. Considering the objectives of these three skills, for omnidirectional locomotion and high platform climbing, the curriculum progresses once the robot successfully maintains velocity tracking for a period of time, and regresses if the robot’s traveled distance is far below the theoretical distance. For fall recovery, the curriculum progresses when the robot is within a certain range of the standard standing posture, and regresses otherwise. The
corresponding terrains are as follows with the terrain level $l\in[1,20]$: 
\begin{itemize}
\item Stairs: with step width of $0.3$ m and step height of $0.05+0.18\times l/20$ m;
\item Slope: with slope of $0.5\times l/20\times 100\%$;
\item Discretized terrain: with 30 rectangular bumps on the ground whose widths range from 1 m to 2 m and heights given by $0.05+0.17\times l/20$ m; 
\item Rough terrain: with height sampled uniformly from $[0.04,0.12]$ m;

\item Pit: a rectangular pit on the ground with dimensions 2 m $\times$ 4 m (length $\times$ width) and depth $0.05+l/20$ m.
\end{itemize}

During Training S2, all environments are treated equally, with the expectation that the robots will learn to select skills on their own to continuously track the velocity commands.

\begin{table}[htbp]
\centering

\captionsetup{font=footnotesize}
\caption{Randomized Simulation Parameters}
\label{tab:domainrand}
\renewcommand\arraystretch{1.5}
\begin{center}
\begin{tabular}{c|c}
\hline
Parameter & Randomization Range \\ 
\hline
Static Friction & $[0.6,1.0]$\\
\hline
Dynamic Friction & $[0.4,0.8]$\\
\hline
Base Mass Bias & $[-1.0,3.0]$ kg \\
\hline
External Force & Every 2-3 s at $[-10.0,10.0]$ N \\
\hline
Push Robots & \makecell[c]{Every 8-12 s at $[-1.0,1.0]$ m/s \\ on $x$ or $y$ direction}\\
\hline
Motor Gain Multiplier & $[0.8,1.2]$ \\
\bottomrule
\end{tabular}
\end{center}

\end{table}
\subsubsection{Domain Randomizations}
To improve the robustness of our proposed method, and to ensure smooth sim-to-real transfer, we widely implement domain randomizations during training. TABLE \ref{tab:domainrand} lists the
randomized variables and their uniformly sampled ranges.

\begin{figure*}
\includegraphics[width=1\textwidth]{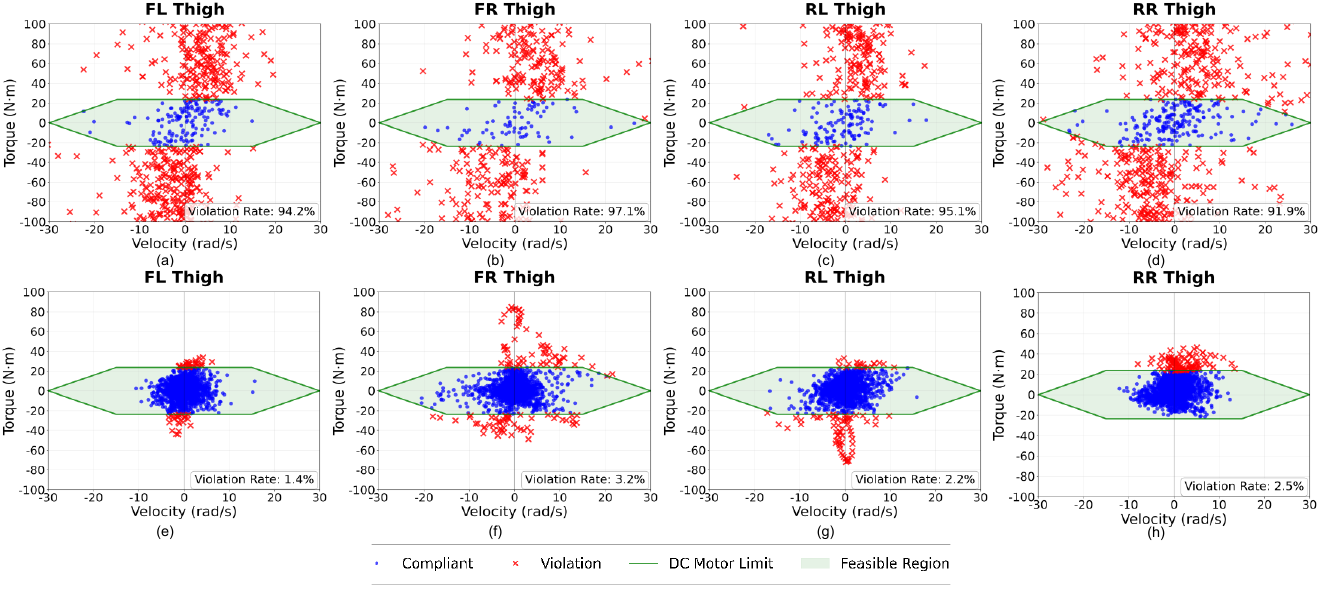}
\captionsetup{font=footnotesize}
\caption{Torque–velocity distributions of thigh joints across diverse terrains. \textcolor{customgreen1}{Green area} indicates compliance with DC-motor constraints. \textcolor{blue}{Blue dots} are motor states within the \textcolor{customgreen1}{Green area}, while \textcolor{red}{red crosses} are not. (a)–(d) show results without DC-motor constraints, while (e)–(h) show constrained results.}
\label{fig:DC-MOTOR-COMPARE}
\end{figure*}
\subsection{Comparison and Ablation Experiments}
\subsubsection{Baseline Comparison and State Estimator Ablation}
We first compare our method with several baselines and ablations on the separate skill performance as follows: 
\begin{itemize}
    \item MUJICA (Ours): Training with all modules. 
    \item MUJICA w/o Velocity: Training without base linear velocity estimation.
    \item MUJICA w/o Wheel Height: Training without wheel-ground distance estimation.
    \item MUJICA w/o Collision: Training without robot segment collision estimation.
    \item DreamWaQ\cite{nahrendra2023dreamwaq} + P3O\cite{zhang2022penalized}: A framework trained with constraints, which explicitly estimates base linear velocity and implicitly infers privileged state using the context-aided VAE estimator network.
    \item Vanilla PPO\cite{schulman2017proximal}: Training with only proprioception.
\end{itemize}

All the above methods are trained under the same terrain curriculum and reward function. To assess policy effectiveness and robustness, 5 representative tasks
are chosen, and 10 increasing difficulty levels are designed, with level gaps doubled relative to training.  Success for omnidirectional locomotion is maintaining a 1 m/s target velocity for 20 s; for recovery, achieving a stable standing pose similar to the nominal posture in 4 s; and for high platform climbing, climbing onto the platform within 3 s and maintaining stability. All
methods are tested with 4 random seeds and mean success rates are reported
alongside the standard deviation.

The results are demonstrated in Fig. \ref{fig:compare}. Firstly, all methods except vanilla PPO are able to recover from slopes at all difficulties since rolling allows the robot to exploit external forces on smooth slope surfaces.
In terms of overall performance, our MUJICA framework utilizing base velocity, wheel-ground distance, and collision estimations consistently outperforms all baselines and ablations across all tasks. Notably, the performance gap widens significantly at higher difficulty levels, demonstrating the effectiveness of our proposed architecture in enhancing robustness. The ablation results indicate that removing any of the three estimation components leads to a noticeable decline in performance, highlighting that accurate evaluation of the surrounding environment requires the joint prediction of all three elements. Moreover, omitting velocity estimation is particularly detrimental for non-recovery tasks, highlighting its importance for adaptation to dynamic environments.

\begin{figure}[htbp]
  \centering
  \captionsetup{font=footnotesize}
\includegraphics[width=\linewidth]{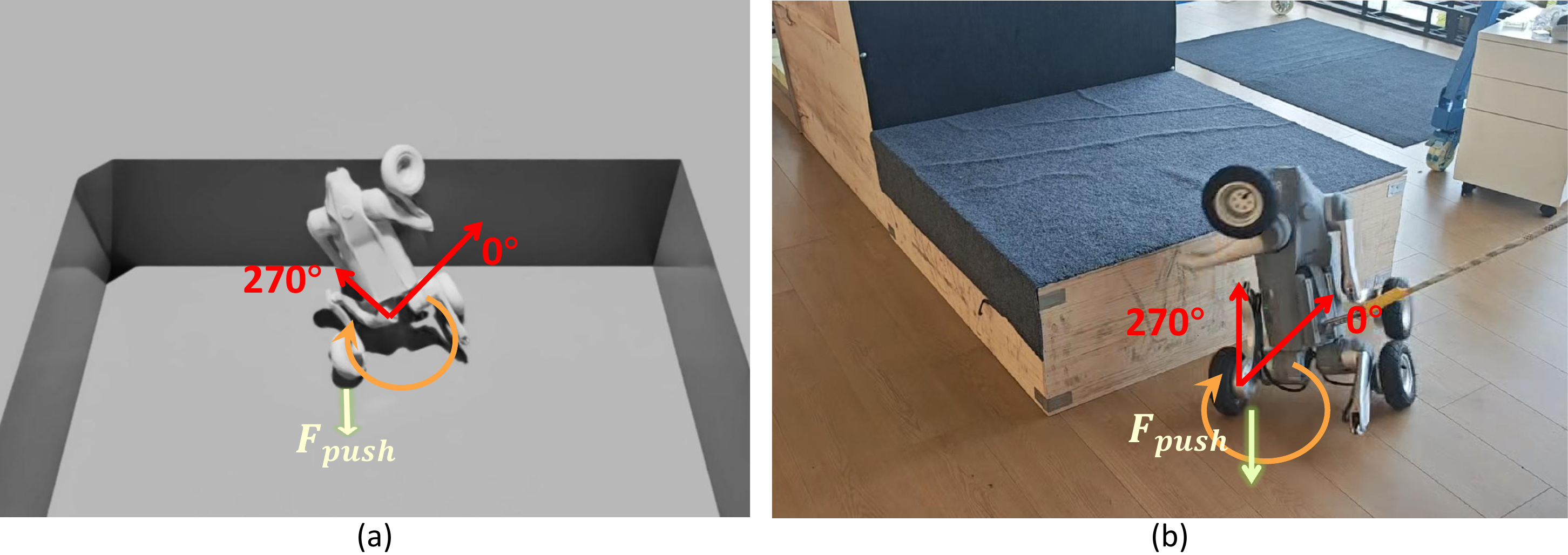}
  \caption{High platform climbing without DC-Motor constraints. The rear thigh joint is at joint limit and will generate large torque to jump highly. (a) In simulation (b) In real world.}
  \label{fig:DC_SIM_REAL}
\end{figure}
Compared to DreamWaQ\cite{nahrendra2023dreamwaq} + P3O\cite{zhang2022penalized}, which also employs an estimation framework with constraints, our method demonstrates superior performance. This suggests that our specific choice of estimation targets and network architecture more effectively captures the necessary information for diverse tasks in wheeled-legged robots. It should be noted that, due to the remarkable mobility of wheeled-legged robots, even policies trained with the simplest PPO baseline algorithm can successfully accomplish the slope-traversing task. However, PPO fails to learn any fall recovery skills due to the absence of state estimation.

\begin{figure*}[htbp]
    
    \centering
    \captionsetup{font=footnotesize,singlelinecheck=false,justification=raggedright}
    \includegraphics[width=\linewidth]{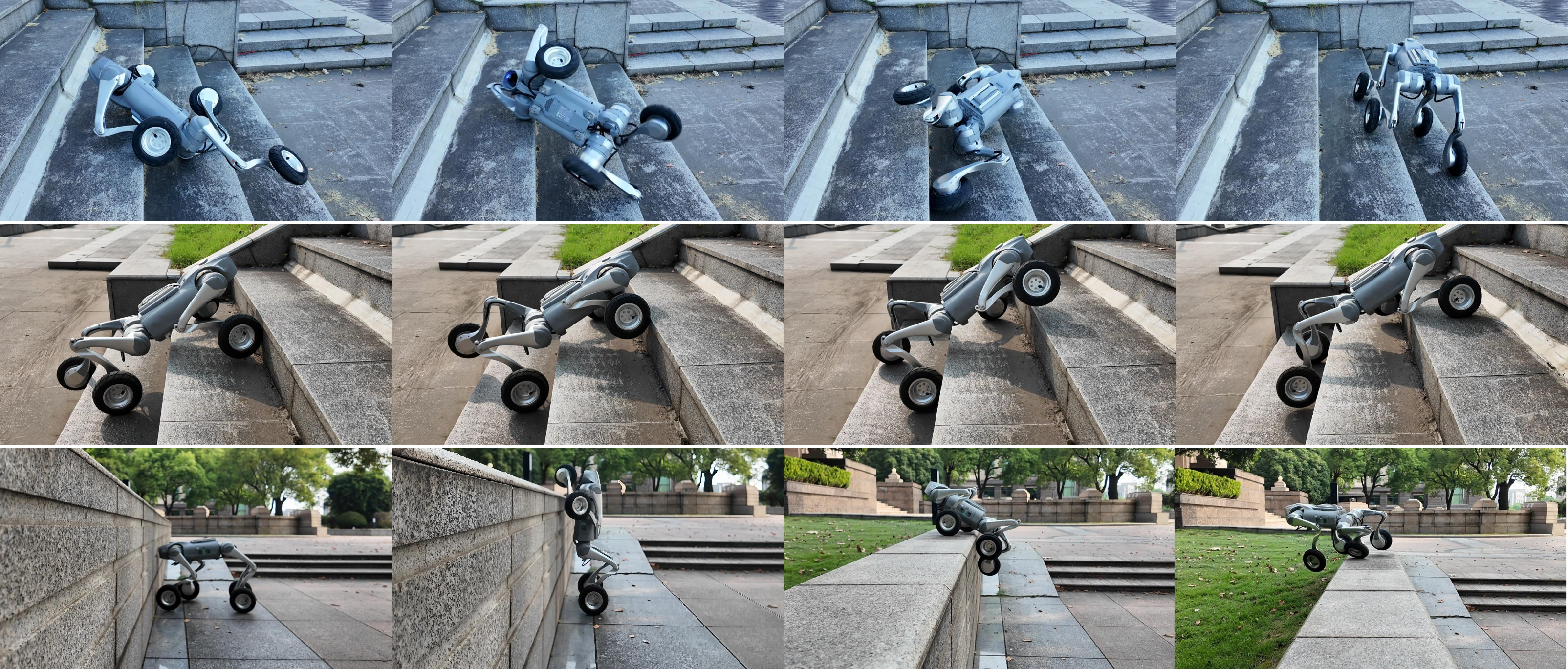}
    \caption{Real-world evaluations of individual skills. \textbf{Top row}: Fall recovery on stairs. \textbf{Middle row}: Ascending 20-cm-high stairs. \textbf{Bottom row}: Climbing an 80-cm-high platform.}
    \label{fig:real_pic}
\end{figure*}

\subsubsection{DC-Motor Constraint Ablation}
To further investigate the role of the DC-motor constraint, we analyze the torque–velocity distributions of the thigh joints across diverse terrains in simulation. As shown in Fig. \ref{fig:DC-MOTOR-COMPARE}, in the unconstrained case (Figs. \ref{fig:DC-MOTOR-COMPARE}a–d), the joints exhibit severe violations, with over 90\% of the samples lying outside the feasible motor region. For clarity, the plots are truncated to ±100 N·m, although many outliers with higher magnitudes exist. Moreover, without the DC-motor constraint, the joints undergo persistent oscillations, and during high platform climbing, the rear thigh joints frequently hit their positive joint limits. At this limit, large torques are generated when pushing against the ground, which directly triggers motor faults on the real robot (as shown in Fig. \ref{fig:DC_SIM_REAL}). In contrast, with the DC-motor constraint enforced (Figs. \ref{fig:DC-MOTOR-COMPARE}e–h), our method effectively keeps the distributions within the feasible region, reduces violation rates to below 3.5\%, and ensures stable, hardware-executable joint behaviors. It's worth noting that the dense cluster of points at zero velocity mainly results from impacts when the robot jumps from outside into the pit, which impose large impulses on the thigh joints.

\begin{table}[htbp] 
\centering 
\captionsetup{font=footnotesize}
\caption{Stage-wise success rate on the sequential terrains}
\label{tab:seq_selector} 
\begin{tabular}{c c c c} 
\toprule Method & \makecell[c]{Stair\\ Recovery} & \makecell[c]{Stair\\ Climbing} & \makecell[c]{High Platform \\Climbing} \\ 
\midrule \makecell[c]{w/o indicator\\ (shared reward)} & \XSolidBold & \XSolidBold & \XSolidBold \\ \hline
\makecell[c]{w/o indicator \\(separate reward)} & 72\% & 72\% & 38\% \\ \hline
\makecell[c]{with skill selector \\(ours)} & \textbf{95\%} & \textbf{95\%} & \textbf{91\%} \\ 
\bottomrule 
\end{tabular} 
\end{table}
\subsection{Evaluation of Skill Selector}
To validate whether the high-level skill selector can autonomously switch skills to handle complex environments, we evaluate three variants on a single episode that chains three tasks at middle difficulty (i.e., difficulty 5 in 10) in order: (i) stair recovery, (ii) stair climbing, and (iii) high platform climbing. Each run must complete all three in sequence to be counted as success. We perform 100 episodes with identical terrains and manually provided commands. Failure is defined as base collision during stair climbing or high platform climbing, or if the task objective cannot be achieved within 3 seconds after  adjusting the command.

We compare our method with the following two baselines:
\begin{itemize}
    \item w/o indicator (shared reward): Training without the skill indicator and with unified rewards across all tasks.
    \item w/o indicator (separate reward): Training without the skill indicator and with reward terms differing by tasks.
\end{itemize}

The results are presented in TABLE \ref{tab:seq_selector}. Policies without a skill indicator may confuse behaviors across tasks, thereby impairing single-task performance. In particular, the unified-reward training scheme often collapses (i.e., fails to learn skills) ,  while the separated-reward scheme still produces lower success rates on medium-difficulty tasks compared to the proposed hierarchical control architecture. This highlights that reward engineering alone is insufficient for multi-skill integration. Without an explicit indicator, the policy struggles to disentangle task-specific strategies and generalize across heterogeneous objectives. In contrast, incorporating the skill indicator and the high-level skill selector enables clear task differentiation, stabilizes training, and ultimately supports reliable sequential execution in complex scenarios.

\subsection{Real-World Experiments}

To validate MUJICA's zero-shot sim-to-real performance, we deploy the policies on the Unitree Go2-W robot in the real world. All onboard computations are performed on an NVIDIA Jetson Orin NX. These experiments aim to assess the robustness of individual skills and the effectiveness of the skill selector in autonomous multi-skill execution.

\subsubsection{Individual Skill Validation} 
We first evaluate each skill under ordinary and challenging conditions. In stair recovery (Fig. \ref{fig:real_pic} Top Row), the robot successfully recovers from an upside-down state on 30° stairs, using calf strikes to generate torque and rotate the base, demonstrating robustness on uneven terrains at random joint positions. In stair climbing (Fig. \ref{fig:real_pic} Middle Row), it steadily ascends stairs with varying riser heights and irregular edges. For high platform climbing, the robot clears an 80-cm-high outdoor platform (Fig. \ref{fig:real_pic} Bottom Row) and achieves a challenging 1-m-high indoor platform (Fig. \ref{fig:head_pic} Bottom Row), approaching its mechanical and actuation limit. The maneuver requires squatting the hind thigh joints to store energy, extending forelegs to hook onto the box, and lifting the hind legs in coordination, which demonstrates full wheel–leg synergy. To our knowledge, no previous work has enabled a wheeled-legged robot to accomplish this extreme task. Across all tasks, the motor protection mechanism—which will be triggered by excessive current or overheating—remains inactive, showing that the algorithm exploits the full potential of wheeled-legged locomotion.

\subsubsection{Multi-Skill Transition Tests} 
We then evaluate MUJICA’s ability to autonomously integrate multiple skills during continuous real-world missions (Fig. \ref{fig:head_pic} Top Row). In one demonstration, the robot starts from a random fallen posture, executes recovery, climbs a stair and a ramp, and finally climbs a 60-cm-high platform. The high-level skill selector seamlessly switches policies based on proprioceptions without human intervention: detecting the fallen posture triggers recovery, resuming the upright posture switches to omnidirectional moving, and encountering the box activates the high-platform-climbing skill. Full demonstrations of real-world experiments are available in the supplementary video.

\section{CONCLUSIONS}
In this work, we introduce MUJICA, a unified control architecture that enables wheeled-legged robots to acquire and integrate multiple challenging locomotion skills within a single proprioceptive policy. By incorporating a skill selector and a safety-aware learning framework with precise DC-motor constraints, MUJICA achieves smooth transitions across diverse tasks such as omnidirectional locomotion, high platform climbing, and fall recovery. Extensive experiments demonstrate that the proposed approach not only improves robustness compared to baseline and ablated variants, but also achieves reliable zero-shot sim-to-real transfer. These results highlight MUJICA's potential to advance the autonomy and adaptability of wheeled-legged robots operating in complex real-world environments. Future work will explore extending the framework to a broader set of locomotion skills and enabling adaptive coordination on unstructured terrains.
\bibliographystyle{IEEEtran}
\bibliography{mylib}

\end{document}